\documentclass{article} 
\usepackage[final]{colm2026_conference}

\usepackage{microtype}
\usepackage{hyperref}
\usepackage{url}
\usepackage{booktabs}
\usepackage{multirow}
\usepackage{xspace}
\usepackage{graphicx}

\usepackage{lineno}

\definecolor{darkblue}{rgb}{0, 0, 0.5}
\hypersetup{colorlinks=true, citecolor=darkblue, linkcolor=darkblue, urlcolor=darkblue}

\newcommand{\bench}{\textsc{MUDDLE}\xspace}

\title{MUDDLE: Measuring Understanding of Documents under Distractor and Length Effects}

\author{
\normalfont
\begin{tabular}{c}
\textbf{Jason Luo}$^{*}$, \enspace \textbf{Saibilila Abudukelimu}$^{*}$, \enspace \textbf{Judy Song}, \enspace \textbf{Andrew Feng}, \\
\textbf{Shivank Garg}, \enspace \textbf{Vasu Sharma}, \enspace \textbf{Kevin Zhu}
\end{tabular}
}

\begin{document}

\ifcolmsubmission
\linenumbers
\fi

\maketitle

\begin{abstract}
Document question-answering systems increasingly answer questions over collections of
retrieved documents rather than one clean source, so robustness to distracting context
matters as much as reading ability. When such systems fail, it is often unclear whether the
context was too long or the distractors were too close to the topic, because prior
work tends to conflate these two effects. We present \bench, a controlled benchmark
that separates them. \bench\ uses 270 human-annotated questions, each tied to a single
source document, and instantiates every question in five conditions: the source alone,
the source with two or four topically similar hard negatives, and the source with two
or four random distractors. The random distractors are matched to the hard negatives
in length and provenance, so an accuracy gap between the two arms reflects topical
similarity rather than length. All five conditions are rendered in markdown, page images, and raw PDF, but the
distractor sweep reported here is run in markdown, since a source plus its distractors
exceeds current image and PDF input limits. We score answers with an LLM judge across three model
families. In the complete markdown sweep, hard negatives lower accuracy more than
length-matched random documents at both context sizes for gpt-5-mini, while random
documents stay near the no-distractor baseline. The effect is small but directionally
consistent, and for gpt-5-mini hard negatives significantly underperform length-matched
random distractors when pooled across context sizes. We
release the data and evaluation code for a reproducible study of context degradation.
\end{abstract}

\section{Introduction}
\label{sec:introduction}

Large vision-language models are increasingly used not to read a single, clean
document but to work over whole collections of documents pulled on demand. An
analyst's question is answered by retrieving reports, papers, and filings; an agent
grounds a claim by reading whatever a search turns up. Two trends drive this. First,
context windows have grown by orders of magnitude, yet models use that capacity
unevenly and degrade as inputs lengthen \citep{hsieh2024ruler}. This tempts systems to
drop an entire retrieved bundle into one prompt rather than curating it first. Second,
retrieval-augmented generation is now the standard way to ground outputs in external
knowledge \citep{lewis2020rag}. The retrieved set is rarely clean: corpora are messy
and retrievers are imperfect, so the one document that answers the question often
arrives buried among plausible look-alikes that do not. Whether a model can still find
that document and reason over it while ignoring the rest is a core property of any
document-grounded system, not an edge case.

Two lines of work speak to this, but neither isolates the failure we care about. One
studies how models use long contexts and finds that they use them unevenly: accuracy
depends on where the evidence sits and declines as the context grows
\citep{liu2024lost}. The other studies retrieval noise and shows that irrelevant or
misleading passages pull answers off course, though carefully placed random documents
can sometimes help \citep{shi2023distracted, cuconasu2024power}. Both rely on short,
text-only, open-domain passages, where a distraction is a sentence or a paragraph
rather than a full document with its own layout. Multimodal long-document benchmarks
close part of the gap. MMLongBench-Doc \citep{ma2024mmlongbench}, for instance, shows
that reading even one lengthy PDF is hard, with the best-performing model they test
(GPT-4o) reaching an F1 of only $44.9\%$. But these benchmarks judge the source
document on its own. They never place it among the related-but-wrong documents a
retriever would surface, so a practitioner cannot tell whether a failure comes from
the length of the context or from distractors that look like the answer. That
distinction decides whether to curate the retrieved set or simply truncate it, and it
is exactly what existing benchmarks leave unmeasured.

Our starting point is that the failure mode worth isolating is not context length on
its own, but semantic adjacency: documents that share a topic and vocabulary with the
source, and so look as though they answer the question, yet do not actually contain the
answer. We call these documents \emph{hard negatives}, borrowing the term from dense
retrieval, where hard negatives are the topically close but non-relevant passages a
retriever is most likely to return \citep{karpukhin2020dpr}. We expect hard negatives
to be more corrosive than the same amount of random text. The form a document takes
when it reaches the model, whether a native PDF, text extracted to markdown, or
rendered page images, also exposes its layout differently, so we treat input modality
as an explicit axis. These considerations lead to four research questions that organize
the paper. \textbf{(RQ1) Topical similarity:} at a matched context length, do hard
negatives degrade answer accuracy more than random distractors? \textbf{(RQ2) Length:}
does adding more distractor documents ($k{=}0 \rightarrow 2 \rightarrow 4$) degrade
accuracy, independent of distractor type? \textbf{(RQ3) Modality, a baseline
check rather than a hypothesis test:} on the source-only condition, does answer accuracy
differ when the same document is presented as markdown, page images, or a raw PDF? \textbf{(RQ4) Model:} do these
effects hold across models from different families? We answer RQ1, RQ2, and RQ4 in
markdown, where the full distractor sweep is servable, and treat the distractor sweep
in image and PDF as an extension that our released code supports (Section~\ref{sec:limitations}). Our distractor findings are
therefore text-based, and we do not claim the same margin holds for page images or PDF.

To answer these questions we built \bench\ on top of MMLongBench-Doc. We start from
270 human-annotated questions over 85 source documents. For each question we assemble
five context conditions that hold the source fixed and change only what surrounds it:
a \emph{control} with the source alone; \emph{HN-2} and \emph{HN-4}, which add two and
four hard negatives; and \emph{Random-2} and \emph{Random-4}, which add two and four
random distractors. The hard negatives are curated to be on topic yet free of the
answer, and the random distractors are drawn so that the two arms differ only in topical
relevance (Section~\ref{sec:random}). Every condition
is rendered in markdown, page images, and raw PDF and scored with an LLM judge for
semantic correctness. The design separates three factors that earlier work tends to mix
together: whether extra context is present, how many distractors there are, and what
kind of distractor it is.

Our contributions are:
\begin{itemize}
  \item A controlled, multi-document evaluation suite that varies distractor type and
  distractor count over long, real-world documents, holding the source document fixed and
  matching context length between distractor types. The same conditions are rendered in
  three input modalities (markdown, page images, and raw PDF); the distractor sweep is
  evaluated in markdown, and the image and PDF renderings support a source-only modality
  comparison.%
  \footnote{Data and code: \url{https://github.com/luoojason/muddle}.}
  \item A hard-negative curation pipeline. We retrieve real documents by search, filter
  them with a model that keeps topical look-alikes while rejecting any that may contain
  the answer, and verify the result by hand. Each hard negative is matched in length to
  the random distractor pool, and each question is checked to be answerable only from its
  source document.
  \item An evaluation protocol and an empirical finding. Under paired, same-question
  comparison on the complete five-condition markdown sweep, topical adjacency degrades
  accuracy beyond length alone for gpt-5-mini, significant when pooled across context
  sizes, while the effect stays small and directional across the three families. Our released code extends the
  protocol to image and PDF distractor conditions once they can be served at scale.
\end{itemize}

\section{Related Work}
\label{sec:related}
\subsection{Long context and modality}
Two properties of modern models motivate our setup. First, models use long contexts
unevenly: accuracy is highest when the relevant evidence sits near the start or end of the
input and lowest in the middle, and it declines as the context grows \citep{liu2024lost},
with stress tests confirming that nominal context windows overstate usable capacity and
most models degrade before their advertised limit \citep{hsieh2024ruler, bai2024longbench}.
We hold source placement fixed and seeded (Section~\ref{sec:benchmark}) so these positional
effects are controlled rather than measured, isolating distractor type at matched length.
Second, the documents a model reads are increasingly multimodal: document question
answering is now posed over rendered pages rather than clean text, beginning with
single-page settings such as DocVQA \citep{mathew2021docvqa}, and the form a document takes
(extracted markdown, page images, or native PDF) exposes layout differently, so we treat
modality as an explicit axis rather than a fixed preprocessing choice.

\subsection{Hard negatives and retrieval noise}
Retrieval-augmented generation grounds model outputs in retrieved documents
\citep{lewis2020rag}, but the retrieved set is rarely clean. Irrelevant or misleading
context lowers answer quality: large language models are easily distracted by irrelevant
material \citep{shi2023distracted}, and how much retrieved noise hurts depends on its
relationship to the query \citep{yoran2024ragrobust}, to the point that some off-topic
documents can even help \citep{cuconasu2024power}. This line of work studies short, text-only,
open-domain passages; we scale the distractor to a full multimodal document and separate
topical from random noise explicitly.

In dense retrieval, hard negatives are mined to train the retriever, first from a lexical
retriever \citep{karpukhin2020dpr} and later globally from the corpus \citep{xiong2021ance}.
We instead import them as an evaluation condition, testing whether a model can still find
and use the one relevant document when it is surrounded by plausible look-alikes.

\subsection{Recent benchmarks}
A growing set of benchmarks evaluates document understanding at length and across
modalities, and our design responds to a gap they share. MMLongBench-Doc poses
human-annotated questions over long PDFs and shows single-document reading is hard even
for strong models, though its renderings can be lossy \citep{ma2024mmlongbench}. DUDE
covers multi-page, visually rich documents and likewise finds a large multi-page gap
\citep{vanlandeghem2023dude}. LongDocURL adds reasoning and locating tasks over long multimodal documents and evaluates
parsed-text and page-image inputs as separate modes, but each question is answered over a
single document rather than among distractors \citep{deng2025longdocurl}. UniDoc-Bench
assembles a multimodal retrieval benchmark of 1{,}600 question-answer pairs comparing text,
image, and fused retrieval, but it varies modality and retrieval strategy rather than
isolating context length or topical-distractor type \citep{peng2025unidocbench}. IRPapers,
for scientific retrieval, evaluates image- and text-based retrieval as separate single-modality conditions \citep{shorten2026irpapers}. Across these, none places the target document among the
related-but-wrong documents a retriever would surface while holding modality and context
length apart. \bench\ is built for exactly that gap.

\section{Benchmark Construction}
\label{sec:benchmark}
\begin{figure}[t]
\centering
\includegraphics[width=0.85\linewidth]{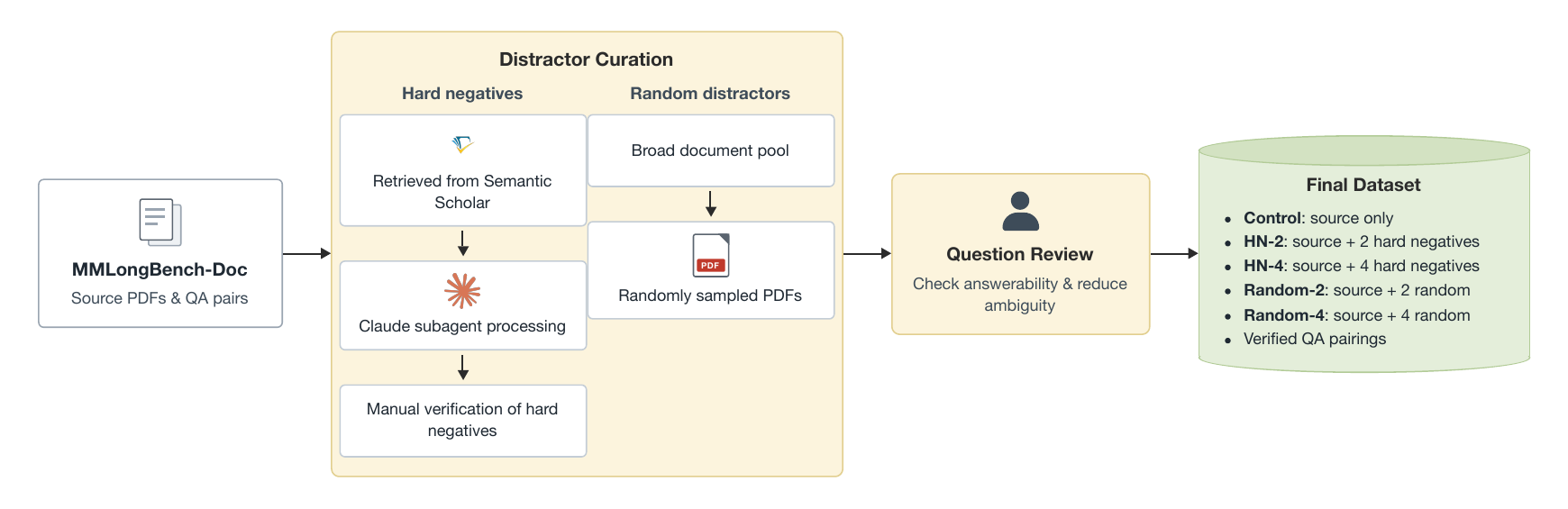}
\caption{Construction of \bench. Human-annotated questions over long PDFs from
MMLongBench-Doc \citep{ma2024mmlongbench} are paired with curated hard negatives
(retrieved by search, filtered by model-assisted review, verified by hand, and held to
a 10 to 40 page band) and with random distractors drawn from the hard negatives of
\emph{other} questions (Section~\ref{sec:random}). Each (question, context) cell is
rendered in three modalities (markdown, page images, and native PDF) and scored with an
LLM judge.}
\label{fig:construction}
\end{figure}

Figure~\ref{fig:construction} summarizes the construction of \bench, from source
questions through distractor curation to multimodal rendering and scoring.

\subsection{Source questions}
\bench\ is built on MMLongBench-Doc \citep{ma2024mmlongbench}, a benchmark for question
answering over long, multimodal PDFs. We keep questions answerable from text and tables, not from graphs (e.g.\ a pie chart), since the markdown rendering drops figures that the page images preserve. The resulting
set has 270 questions over 85 source documents (1 to 17 per document), with every question
and reference answer tracing to the original human annotations rather than model
generation. We labeled each question by task type, giving a roughly even split between
recall questions (retrieve a stated fact) and reasoning questions (combine information
across the document).

\subsection{Context conditions}
For each question we build five context conditions that hold the source document fixed
and vary only what surrounds it: a \textbf{control} with the
source alone; \textbf{HN-2} and \textbf{HN-4}, adding two and four \emph{hard negatives}
(documents topically similar to the source); and \textbf{Random-2} and \textbf{Random-4},
adding two and four \emph{random} distractors drawn without regard to topic. Crossing 270
questions with five conditions gives 1{,}350 (question, context) cells per modality.

\subsection{Hard-negative curation}
We assign each source document four hard negatives, mined specifically for that source. Hard negatives are sourced by search over signals from the source
document, the question, the answer, and the document topic. A model-assisted review then
keeps documents similar to the source while rejecting those likely to contain the
answer, and a human checks the result. Every shipped hard negative passes two automatic
gates: a token-overlap (Jaccard) below $0.45$ against the source, so distractors are
topical but not near-duplicates, and a word-boundary answer-leak check.
To keep context length from confounding the topical-similarity comparison, every hard
negative is constrained to a band of 10 to 40 pages. If a source had more than four mined hard negatives, two were chosen at random for HN-2, then two more were drawn from the rest and added to them for HN-4. Appendix~\ref{app:construction} reports the five conditions in full
(Table~\ref{tab:conditions}) and details the manual-review axes, length statistics, and
provenance.

\subsection{Hard-negative retrieval and filtering}
\label{sec:retrieval}

\paragraph{Candidate retrieval.} For each source document we retrieve a pool of distractor
candidates by topical search. Academic source documents are matched against the Semantic
Scholar Graph and paper-recommendation APIs using a title extracted from the source; we
keep each candidate's title, abstract, year, and identifiers and take the top eight by
retrieval rank. Non-academic sources, such as product manuals, financial reports, and
court filings, are matched instead by keyword web search restricted to PDF files, with
keywords drawn from a TF-IDF analysis of the source text and its questions; we keep
candidates whose TF-IDF cosine similarity to the source falls in a moderate band (about
$0.30$ to $0.85$), so a candidate is related to the source without being a near-duplicate.
Candidates already in the benchmark, missing an abstract, or duplicating one another are
discarded.

\paragraph{Model-assisted filtering.} Answer leakage is checked with Claude Haiku 4.5 in
two steps: the model answers the question from only the candidate's title and abstract,
replying that it cannot when the text is insufficient; if it instead answers, a second
call judges whether that answer matches the reference, and any candidate yielding a match
is removed as contaminated. Surviving candidates are then rated by a Claude reviewer on
three axes, whether the candidate shares the subject of the source, whether it leaks the
answer, and whether its document type fits, and a fixed rule maps these ratings to a good,
weak, or bad verdict. Candidates rated weak or bad are replaced from a pool of spare
candidates, and the survivors pass to the manual review in Appendix~\ref{app:construction}.

\subsection{Random distractors}
\label{sec:random}
Random distractors are the hard negatives belonging to \emph{other} questions. For a
given question we draw from the 339-document pool, excluding that question's own hard
negatives and source, and sample four documents with a per-question seed. Random-2 is
the first two of that draw, so it nests inside Random-4. Because each random distractor
is itself a hard negative of some other question, the random arm shares the provenance
and the 10 to 40 page length distribution of the hard-negative arm. The two arms differ
only in topical relatedness to the current source. At matched $k$ their page counts are
close (median 84 against 86 pages at $k{=}2$), so an accuracy gap between hard negatives
and random distractors at the same $k$ is attributable mainly to topical similarity
rather than length. A random distractor can, by chance, be topically related to the
source; we treat such accidental hard negatives as a known source of noise.

\subsection{Source placement}
For conditions with at least one distractor, the source is inserted at a seeded,
reproducible position keyed on the cell identifier, identical across all models and
modalities. This holds positional effects \citep{liu2024lost} constant across our
comparisons.

\subsection{Modality renderings}
The same documents are presented in three forms.
\textbf{Markdown:} each PDF is converted to markdown with MarkItDown~\citep{markitdown}
and documents are concatenated under per-document headers.
\textbf{Image:} each PDF page is rendered to a high-resolution PNG by a headless-browser
pipeline (Playwright~\citep{playwright} driving pdf.js~\citep{pdfjs} at viewport scale
$2.0$, one image per page). Because full-resolution PNGs would exceed provider
request-size limits, the page images are downgraded at send time, re-encoded as
downscaled JPEGs (Appendix~\ref{app:construction} reports the resolutions). Each cell's
rendering is held identical across all models, and we treat the resulting resolution
variation as a known confound (Section~\ref{sec:limitations}). For the OpenAI model we set the API \texttt{detail}
parameter to \texttt{high}, which the Gemini and Grok APIs do not expose.
\textbf{PDF:} the raw PDF files are passed directly to the model as file blocks, except
for \texttt{grok-4.3}, which has no native PDF input and instead receives OpenRouter text
extraction or Mistral OCR, so its PDF column is not comparable to the other two.
Appendix~\ref{app:stats} reports per-cell size statistics for each modality and condition.

\section{Experimental Setup}
\label{sec:setup}

\subsection{Models}
We evaluate three models from different families: \texttt{gpt-5-mini},
\texttt{gemini-3.5-flash}, and \texttt{grok-4.3}.
\texttt{gpt-5-mini} and \texttt{gemini-3.5-flash} run at a fixed medium reasoning
effort. Providers differ in how they expose this control, so it is passed where
supported and omitted otherwise, and \texttt{grok-4.3} runs without an explicit effort
parameter. Requests are issued through a single client library (litellm) with a uniform
message format. Routing is by provider and request size: \texttt{gpt-5-mini} runs on
Azure OpenAI for smaller cells and on OpenRouter for cells above the per-request image
limit, \texttt{gemini-3.5-flash} runs on OpenRouter, and \texttt{grok-4.3} runs on
Azure AI Foundry.

\subsection{Prompt and inference}
Each query is a single user message containing a short instruction, the question,
and the documents in their placed order. The instruction is held fixed up to the
phrase naming the document format. For markdown it reads: ``You are given one or more
documents (converted to markdown). Using ONLY their content, answer the question
with a short, direct answer and nothing else. If the documents do not contain the
answer, reply `Not answerable'.'' The image and PDF instructions are identical
except for ``documents as page images'' and ``PDF documents''. In the markdown and
image conditions, documents are separated by per-document headers; in the PDF
condition each document is a separate file block identified by filename. The full
prompts are reproduced in Appendix~\ref{app:prompts}.

\subsection{Metrics}
Our primary metric is an LLM judge \citep{zheng2023judging, liu2023geval} that scores
semantic correctness. The judge
(\texttt{gpt-5.4-nano}, reasoning effort minimal) receives the question, the
reference answer, and the model prediction, and returns a JSON verdict with a
boolean \texttt{correct}, a graded \texttt{score} in $[0,1]$ for partial credit,
and a one-sentence rationale. The judge prompt specifies equivalence rules for
numbers, units, and list order, and marks refusals incorrect; we use the boolean
\texttt{correct} as the headline accuracy. The judge implementation is released with our
evaluation code.
As secondary, fully reproducible metrics we report exact match and token-level F1
over SQuAD-style normalization (lowercasing, article and punctuation removal, and
whitespace collapse) \citep{rajpurkar2016squad}. We report the judge as primary
because answers are short and free-form, so string-overlap metrics understate
correctness, especially for models that answer verbosely. 

\subsection{Reporting protocol}
All distractor-effect comparisons are computed \emph{paired}: for each
comparison we restrict to the questions answered in both arms and average the
per-question difference, rather than comparing raw per-condition averages over
mismatched samples. The effects we study are a few points in size, so pairing is
necessary to read them at all.



\begin{table}[t]
\centering
\small
\begin{tabular}{llccc}
\toprule
\textbf{Model} & \textbf{Condition} & \textbf{Markdown} & \textbf{Image} & \textbf{PDF} \\
\midrule
\multirow{5}{*}{gpt-5-mini}
 & Control  & $0.763$ & $0.804$ & $0.804$ \\
 & HN-2     & $0.737$ & -- & -- \\
 & Random-2 & $0.767$ & -- & -- \\
 & HN-4     & $0.707$ & -- & -- \\
 & Random-4 & $0.748$ & -- & -- \\
\midrule
\multirow{5}{*}{gemini-3.5-flash}
 & Control  & $0.733$ & $0.789$ & $0.778$ \\
 & HN-2     & $0.711$ & -- & -- \\
 & Random-2 & $0.722$ & -- & -- \\
 & HN-4     & $0.726$ & -- & -- \\
 & Random-4 & $0.711$ & -- & -- \\
\midrule
\multirow{5}{*}{grok-4.3}
 & Control  & $0.744$ & $0.814$ & $0.667$ \\
 & HN-2     & $0.733$ & -- & -- \\
 & Random-2 & $0.726$ & -- & -- \\
 & HN-4     & $0.682$ & -- & -- \\
 & Random-4 & $0.707$ & -- & -- \\
\bottomrule
\end{tabular}
\caption{Judge accuracy by model, modality, and condition (primary metric, higher is
better). Markdown reports all five conditions for all three models. Image and
PDF report the control condition only; their distractor conditions are infeasible at
scale (Section~\ref{sec:limitations}) and are marked --. Per-model F1 and EM are in
Appendix~\ref{app:stats}.}
\label{tab:evaluation_results}
\end{table}

\section{Results and Analysis}
\label{sec:analysis}

We organize the analysis around the four research questions, reading effects from paired,
same-question comparisons (Section~\ref{sec:setup}). All three models are run on all five markdown conditions, plus control runs in image and PDF. We
anchor distractor-effect claims on \texttt{gpt-5-mini} markdown, which has the widest
spread across conditions, so a few-point effect is detectable across the 270 questions; we read the
cross-model and cross-modality picture as directional given the small effect sizes
(Section~\ref{sec:limitations}).

\paragraph{Effect of Topical Similarity.}
At matched context length, hard negatives degrade accuracy more than random distractors.
For \texttt{gpt-5-mini} in markdown, hard negatives score below the length-matched random
arm by $0.030$ at $k{=}2$ and by $0.041$ at $k{=}4$ (both paired), where each gap
averages the per-question difference over the questions answered in both arms; the
per-condition means are in
Table~\ref{tab:evaluation_results}. Because hard-negative and random cells are
length-matched at each $k$ (Section~\ref{sec:benchmark}), this gap is attributable mainly
to topical similarity rather than length. The effect is small, a few points, and grows
modestly from $k{=}2$ to $k{=}4$; pooled across the two context sizes the topical gap is
statistically significant (paired $p{=}0.016$).

\paragraph{Effect of Length.}
Adding more distractors lowers accuracy within each type. The pure length effect is
isolated by the random arm, whose documents add length without deliberate topical overlap
(Section~\ref{sec:random}), and it is weaker than the topical effect. For \texttt{gpt-5-mini} markdown, paired accuracy falls
from $k{=}2$ to $k{=}4$ by $0.019$ in the random arm against $0.030$ in the hard-negative
arm, and the random conditions stay close to control throughout
(Table~\ref{tab:evaluation_results}). Unrelated documents thus cost little as context
grows, while added context bites mainly when the documents are topically adjacent. That
contrast is the central reason to separate distractor type from count.

\paragraph{Effect of Modality.}
As stated in RQ3, modality here is a control baseline rather than a distractor test: the
image and PDF distractor conditions cannot be served (Section~\ref{sec:limitations}), so
the comparison is over the source-only condition. On control, the same questions are read
about equally well across modalities, if anything slightly better from images and native
PDF than from markdown. For \texttt{gpt-5-mini} the control scores are $0.763$ (markdown),
$0.804$ (PDF), and $0.804$ (image), and \texttt{gemini-3.5-flash} shows the same ordering
(Table~\ref{tab:evaluation_results}). Whether \emph{topical degradation}
transfers across modalities is left to an extension that our code supports.

\paragraph{Effect of Model Family.}
The picture varies by model family. \texttt{gpt-5-mini} has the clearest ordering,
with hard negatives below random at both $k$, and its drop from control to HN-4 is
significant (paired $p{=}0.008$), the strongest single effect in our data.
\texttt{gemini-3.5-flash} is markedly
flatter and barely degrades, so it appears robust to topical distractors at $k\le 4$,
consistent with its long-context strength. \texttt{grok-4.3} shows no
gap at $k{=}2$ and degrades only under the heaviest load (control to HN-4, paired
$p{=}0.026$). Where the load is heaviest, the hard-negative effect is significant for the
two models that degrade at all (\texttt{gpt-5-mini} and \texttt{grok-4.3}); the finer
hard-negative-versus-random contrast is clearest for \texttt{gpt-5-mini} and directional
across the panel. For \texttt{grok-4.3}, judge accuracy is
the only trustworthy metric, since its F1 and EM are deflated by verbose answers rather
than by lower correctness (Section~\ref{sec:limitations}). Establishing whether the
topical effect generalizes is the main open question (Section~\ref{sec:limitations}).

\section{Limitations}
\label{sec:limitations}

Our coverage is necessarily uneven. We report the full five-condition sweep only for
markdown; for image and PDF we report the control (source-only) condition and do not
evaluate distractor settings because combining source and distractor documents routinely
exceeds current input-size limits (Appendix~\ref{app:construction}). Consequently, RQ1
and RQ2 are answered using markdown alone, and our distractor findings should be read as
text-based. The length match between the arms is also approximate: a residual token gap
remains between the two pools (Appendix~\ref{app:construction}), and because our accuracy
differences are small, we cannot fully separate topical adjacency from that gap here; a
per-item mixed-effects model with token count as a covariate would settle it. Our
accuracies also come from a single evaluation run per cell, so we present the
hard-negative penalty as directional rather than precise.

Real documents buy realism at some cost in control: unlike synthetic suites such as RULER
\citep{hsieh2024ruler}, they vary in style, complexity, and information density, and our
paired design fixes the source but not the distractors. Curation is human-checked for
topicality and answer leakage, which makes the comparison trustworthy at 270 questions
but bounds its scale; automating that check with a judge ensemble is the most promising
route to a larger version. The model panel is also incomplete: evaluating
\texttt{claude-sonnet-4.6} was beyond the budget of this project and is left for future
work. In addition, \texttt{grok-4.3} produces highly verbose responses, which
substantially depress lexical F1 and EM even when the judge marks the answer correct,
making judge accuracy the most reliable metric for that model
(Appendix~\ref{app:construction}). Our released codebase is designed to support image and
PDF distractor conditions once they become servable, and future work should extend the
evaluation to larger distractor loads (e.g., $k{=}8$) and introduce a genuinely unrelated
distractor arm to provide a less conservative baseline than the current
hard-negative-derived random pool.

\section{Conclusion}
\label{sec:conclusion}

\bench\ separates two effects that long-context and retrieval-noise studies tend to mix:
how far a distractor sits from the source in topic, and how much length it adds. The
design holds the source document fixed, matches distractor length across the
hard-negative and random arms, and renders every condition in markdown, page images, and
PDF. The hard-negative curation pipeline, retrieval followed by model-assisted filtering
and human verification, is what makes the matched comparison possible, since it yields
distractors that are reliably on topic, free of the answer, and length-matched to the
random pool.

On the complete markdown sweep, topical adjacency rather than length drives the
degradation we observe. For \texttt{gpt-5-mini}, hard negatives cost more than
length-matched random documents at both context sizes, while random documents stay near
the no-distractor baseline. The effect is small and model-dependent, and most pronounced
at the larger context size. We read it as a directional result that a stronger design
should be able to resolve.

Read directionally, this suggests that distractor type, not just distractor count, may
matter for how much retrieval noise hurts, and that a benchmark aiming to predict
deployment behavior should control topical adjacency rather than length alone. Confirming
it will take the stronger design we describe (Section~\ref{sec:limitations}).

\section*{Ethics Statement}
The benchmark is built from an existing, publicly available document QA dataset and
from documents retrieved by web search for use as distractors. Questions and answers
trace to existing human annotations. We note that two source documents triggered a
provider content filter and were handled during curation; the released data and
evaluation code are intended for measuring robustness to distracting context.

\bibliography{references}

\begin{thebibliography}{21}
\providecommand{\natexlab}[1]{#1}
\providecommand{\url}[1]{\texttt{#1}}
\expandafter\ifx\csname urlstyle\endcsname\relax
  \providecommand{\doi}[1]{doi: #1}\else
  \providecommand{\doi}{doi: \begingroup \urlstyle{rm}\Url}\fi

\bibitem[Bai et~al.(2024)Bai, Lv, Zhang, Lyu, Tang, Huang, Du, Liu, Zeng, Hou, Dong, Tang, and Li]{bai2024longbench}
Yushi Bai, Xin Lv, Jiajie Zhang, Hongchang Lyu, Jiankai Tang, Zhidian Huang, Zhengxiao Du, Xiao Liu, Aohan Zeng, Lei Hou, Yuxiao Dong, Jie Tang, and Juanzi Li.
\newblock {LongBench}: A bilingual, multitask benchmark for long context understanding.
\newblock In \emph{Proceedings of the 62nd Annual Meeting of the Association for Computational Linguistics (ACL), Volume 1: Long Papers}, pp.\  3119--3137, 2024.

\bibitem[Cuconasu et~al.(2024)Cuconasu, Trappolini, Siciliano, Filice, Campagnano, Maarek, Tonellotto, and Silvestri]{cuconasu2024power}
Florin Cuconasu, Giovanni Trappolini, Federico Siciliano, Simone Filice, Cesare Campagnano, Yoelle Maarek, Nicola Tonellotto, and Fabrizio Silvestri.
\newblock The power of noise: Redefining retrieval for {RAG} systems.
\newblock In \emph{Proceedings of the 47th International ACM SIGIR Conference on Research and Development in Information Retrieval (SIGIR)}, pp.\  719--729, 2024.

\bibitem[Deng et~al.(2025)Deng, Yuan, Bu, Wang, Li, Xu, Li, Gao, Song, Zheng, and Liu]{deng2025longdocurl}
Chao Deng, Jiale Yuan, Pi~Bu, Peijie Wang, Zhong-Zhi Li, Jian Xu, Xiao-Hui Li, Yuan Gao, Jun Song, Bo~Zheng, and Cheng-Lin Liu.
\newblock {LongDocURL}: A comprehensive multimodal long document benchmark integrating understanding, reasoning, and locating.
\newblock In \emph{Proceedings of the 63rd Annual Meeting of the Association for Computational Linguistics (ACL), Volume 1: Long Papers}, pp.\  1135--1159, 2025.

\bibitem[Hsieh et~al.(2024)Hsieh, Sun, Kriman, Acharya, Rekesh, Jia, Zhang, and Ginsburg]{hsieh2024ruler}
Cheng-Ping Hsieh, Simeng Sun, Samuel Kriman, Shantanu Acharya, Dima Rekesh, Fei Jia, Yang Zhang, and Boris Ginsburg.
\newblock {RULER}: What's the real context size of your long-context language models?
\newblock In \emph{Proceedings of the First Conference on Language Modeling (COLM)}, 2024.

\bibitem[Karpukhin et~al.(2020)Karpukhin, O{\u{g}}uz, Min, Lewis, Wu, Edunov, Chen, and Yih]{karpukhin2020dpr}
Vladimir Karpukhin, Barlas O{\u{g}}uz, Sewon Min, Patrick Lewis, Ledell Wu, Sergey Edunov, Danqi Chen, and Wen-tau Yih.
\newblock Dense passage retrieval for open-domain question answering.
\newblock In \emph{Proceedings of the Conference on Empirical Methods in Natural Language Processing (EMNLP)}, pp.\  6769--6781, 2020.

\bibitem[Lewis et~al.(2020)Lewis, Perez, Piktus, Petroni, Karpukhin, Goyal, K{\"u}ttler, Lewis, Yih, Rockt{\"a}schel, Riedel, and Kiela]{lewis2020rag}
Patrick Lewis, Ethan Perez, Aleksandra Piktus, Fabio Petroni, Vladimir Karpukhin, Naman Goyal, Heinrich K{\"u}ttler, Mike Lewis, Wen-tau Yih, Tim Rockt{\"a}schel, Sebastian Riedel, and Douwe Kiela.
\newblock Retrieval-augmented generation for knowledge-intensive {NLP} tasks.
\newblock In \emph{Advances in Neural Information Processing Systems (NeurIPS)}, volume~33, pp.\  9459--9474, 2020.

\bibitem[Liu et~al.(2024)Liu, Lin, Hewitt, Paranjape, Bevilacqua, Petroni, and Liang]{liu2024lost}
Nelson~F. Liu, Kevin Lin, John Hewitt, Ashwin Paranjape, Michele Bevilacqua, Fabio Petroni, and Percy Liang.
\newblock Lost in the middle: How language models use long contexts.
\newblock \emph{Transactions of the Association for Computational Linguistics (TACL)}, 12:\penalty0 157--173, 2024.

\bibitem[Liu et~al.(2023)Liu, Iter, Xu, Wang, Xu, and Zhu]{liu2023geval}
Yang Liu, Dan Iter, Yichong Xu, Shuohang Wang, Ruochen Xu, and Chenguang Zhu.
\newblock {G-Eval}: {NLG} evaluation using {GPT-4} with better human alignment.
\newblock In \emph{Proceedings of the Conference on Empirical Methods in Natural Language Processing (EMNLP)}, pp.\  2511--2522, 2023.

\bibitem[Ma et~al.(2024)Ma, Zang, Chen, Chen, Jiao, Li, Lu, Liu, Ma, Dong, Zhang, Pan, Jiang, Wang, Cao, and Sun]{ma2024mmlongbench}
Yubo Ma, Yuhang Zang, Liangyu Chen, Meiqi Chen, Yizhu Jiao, Xinze Li, Xinyuan Lu, Ziyu Liu, Yan Ma, Xiaoyi Dong, Pan Zhang, Liangming Pan, Yu-Gang Jiang, Jiaqi Wang, Yixin Cao, and Aixin Sun.
\newblock {MMLongBench-Doc}: Benchmarking long-context document understanding with visualizations.
\newblock In \emph{Advances in Neural Information Processing Systems (NeurIPS), Datasets and Benchmarks Track}, volume~37, pp.\  95963--96010, 2024.

\bibitem[Mathew et~al.(2021)Mathew, Karatzas, and Jawahar]{mathew2021docvqa}
Minesh Mathew, Dimosthenis Karatzas, and C.~V. Jawahar.
\newblock {DocVQA}: A dataset for {VQA} on document images.
\newblock In \emph{Proceedings of the IEEE/CVF Winter Conference on Applications of Computer Vision (WACV)}, pp.\  2199--2208, 2021.

\bibitem[{Microsoft}(2024{\natexlab{a}})]{markitdown}
{Microsoft}.
\newblock {MarkItDown}: A tool for converting files to markdown.
\newblock \url{https://github.com/microsoft/markitdown}, 2024{\natexlab{a}}.

\bibitem[{Microsoft}(2024{\natexlab{b}})]{playwright}
{Microsoft}.
\newblock {Playwright}: Fast and reliable end-to-end testing for modern web apps.
\newblock \url{https://github.com/microsoft/playwright}, 2024{\natexlab{b}}.

\bibitem[{Mozilla}(2024)]{pdfjs}
{Mozilla}.
\newblock {PDF.js}: A general-purpose, web standards-based platform for parsing and rendering pdfs.
\newblock \url{https://github.com/mozilla/pdf.js}, 2024.

\bibitem[Peng et~al.(2025)Peng, Qin, Chen, Xu, Xiong, and Wu]{peng2025unidocbench}
Xiangyu Peng, Can Qin, Zeyuan Chen, Ran Xu, Caiming Xiong, and Chien-Sheng Wu.
\newblock {UniDoc-Bench}: A unified benchmark for document-centric multimodal {RAG}.
\newblock arXiv preprint arXiv:2510.03663, 2025.
\newblock URL \url{https://arxiv.org/abs/2510.03663}.

\bibitem[Rajpurkar et~al.(2016)Rajpurkar, Zhang, Lopyrev, and Liang]{rajpurkar2016squad}
Pranav Rajpurkar, Jian Zhang, Konstantin Lopyrev, and Percy Liang.
\newblock {SQuAD}: 100,000+ questions for machine comprehension of text.
\newblock In \emph{Proceedings of the Conference on Empirical Methods in Natural Language Processing (EMNLP)}, pp.\  2383--2392, 2016.

\bibitem[Shi et~al.(2023)Shi, Chen, Misra, Scales, Dohan, Chi, Sch{\"a}rli, and Zhou]{shi2023distracted}
Freda Shi, Xinyun Chen, Kanishka Misra, Nathan Scales, David Dohan, Ed~H. Chi, Nathanael Sch{\"a}rli, and Denny Zhou.
\newblock Large language models can be easily distracted by irrelevant context.
\newblock In \emph{Proceedings of the 40th International Conference on Machine Learning (ICML)}, volume 202 of \emph{Proceedings of Machine Learning Research}, pp.\  31210--31227, 2023.

\bibitem[Shorten et~al.(2026)Shorten, Skaburskas, Jones, Pierse, Esposito, Trengrove, Dilocker, and van Luijt]{shorten2026irpapers}
Connor Shorten, Augustas Skaburskas, Daniel~M. Jones, Charles Pierse, Roberto Esposito, John Trengrove, Etienne Dilocker, and Bob van Luijt.
\newblock {IRPAPERS}: A visual document benchmark for scientific retrieval and question answering.
\newblock arXiv preprint arXiv:2602.17687, 2026.
\newblock URL \url{https://arxiv.org/abs/2602.17687}.

\bibitem[Van~Landeghem et~al.(2023)Van~Landeghem, Powalski, Tito, Jurkiewicz, Blaschko, Borchmann, Coustaty, Moens, Pietruszka, Anckaert, Stanislawek, J{\'o}ziak, and Valveny]{vanlandeghem2023dude}
Jordy Van~Landeghem, Rafal Powalski, Rub{\`e}n Tito, Dawid Jurkiewicz, Matthew~B. Blaschko, Lukasz Borchmann, Micka{\"e}l Coustaty, Sien Moens, Michal Pietruszka, Bertrand Anckaert, Tomasz Stanislawek, Pawel J{\'o}ziak, and Ernest Valveny.
\newblock Document understanding dataset and evaluation ({DUDE}).
\newblock In \emph{Proceedings of the IEEE/CVF International Conference on Computer Vision (ICCV)}, pp.\  19471--19483, 2023.

\bibitem[Xiong et~al.(2021)Xiong, Xiong, Li, Tang, Liu, Bennett, Ahmed, and Overwijk]{xiong2021ance}
Lee Xiong, Chenyan Xiong, Ye~Li, Kwok-Fung Tang, Jialin Liu, Paul~N. Bennett, Junaid Ahmed, and Arnold Overwijk.
\newblock Approximate nearest neighbor negative contrastive learning for dense text retrieval.
\newblock In \emph{The Ninth International Conference on Learning Representations (ICLR)}, 2021.

\bibitem[Yoran et~al.(2024)Yoran, Wolfson, Ram, and Berant]{yoran2024ragrobust}
Ori Yoran, Tomer Wolfson, Ori Ram, and Jonathan Berant.
\newblock Making retrieval-augmented language models robust to irrelevant context.
\newblock In \emph{The Twelfth International Conference on Learning Representations (ICLR)}, 2024.

\bibitem[Zheng et~al.(2023)Zheng, Chiang, Sheng, Zhuang, Wu, Zhuang, Lin, Li, Li, Xing, Zhang, Gonzalez, and Stoica]{zheng2023judging}
Lianmin Zheng, Wei-Lin Chiang, Ying Sheng, Siyuan Zhuang, Zhanghao Wu, Yonghao Zhuang, Zi~Lin, Zhuohan Li, Dacheng Li, Eric~P. Xing, Hao Zhang, Joseph~E. Gonzalez, and Ion Stoica.
\newblock Judging {LLM}-as-a-judge with {MT-Bench} and chatbot arena.
\newblock In \emph{Advances in Neural Information Processing Systems (NeurIPS), Datasets and Benchmarks Track}, 2023.

\end{thebibliography}
\bibliographystyle{colm2026_conference}

\section*{Author Contributions}
Authors marked $^{*}$ (Jason Luo and Saibilila Abudukelimu) contributed equally. Jason Luo designed the benchmark,
built the shared evaluation pipeline, curated the hard negatives, and ran the GPT and Claude
evaluations; Saibilila Abudukelimu built and ran the Gemini evaluations and baselines. Judy
Song and Andrew Feng prepared the page-image and markdown tracks. Shivank Garg advised and
supervised. Vasu Sharma and Kevin Zhu helped with the project reviews and on framing and
writing.

\appendix

\section{Appendix}

\subsection{Benchmark construction details}
\label{app:construction}

\begin{table}[h]
\centering
\small
\begin{tabular}{llcl}
\toprule
\textbf{Condition} & \textbf{Distractor type} & \textbf{$k$ (added docs)} & \textbf{Purpose} \\
\midrule
Control   & none          & 0 & Source-only ceiling \\
HN-2      & hard negative & 2 & Topical similarity, short context \\
Random-2  & random        & 2 & Length-matched baseline for HN-2 \\
HN-4      & hard negative & 4 & Topical similarity, longer context \\
Random-4  & random        & 4 & Length-matched baseline for HN-4 \\
\bottomrule
\end{tabular}
\caption{The five context conditions. Each is instantiated for all 270 questions
and rendered in all three modalities. The source document is identical across
conditions; only the surrounding distractors change.}
\label{tab:conditions}
\end{table}

\paragraph{Hard-negative review.} Manual review recorded a verdict per question across
explicit axes: whether the answer is recoverable from the source, whether any hard
negative leaks the answer, whether hard negatives duplicate one another, whether they
remain on topic, whether the question is answerable from general knowledge alone, and
whether any hard negative is a dead giveaway. Of 274 reviewed items, 232 passed and 42
were flagged for repair. The answer-leak gate uses word-boundary matching that skips
short tokens and bare small numbers to avoid spurious matches.

\paragraph{Length control.} Hard negatives fall in a band of 10 to 40 pages (median 23
pages, standard deviation 8.6). Documents outside the band were replaced by a topically
matched, in-band substitute rather than truncated; 86 such replacements across 56 source
documents are logged with provenance, and we verified that every distractor document
falls inside $[10, 40]$ pages.

\paragraph{Random-distractor sampling.} The per-question draw uses seed
\texttt{randsel\_v2|\{qid\}}, excluding that question's own hard negatives and source;
Random-2 is the first two of the same four-document draw, so it nests inside Random-4.

\paragraph{Source placement.} The insertion position uses a seeded random number
generator keyed on the cell identifier (seed 42), so the position is fixed and
reproducible; the same key is used across all models and all three modalities, so source
placement is identical for a given cell everywhere it appears.

\paragraph{Serving constraints.} Multi-document image and PDF requests exceed provider
limits. Azure OpenAI caps a request at 50 images and routes PDF blocks through the same
pipeline (an effective limit of about 50 pages), which a source plus two to four 10 to 40
page distractors routinely exceeds; larger requests must use OpenRouter, whose gateway
aborts uploads past roughly 15\,MB. Across the suite, 24.7\% of all cells exceed 15\,MB,
concentrated in the distractor conditions (4\% of control, 16\% of $k{=}2$, 43\% of
$k{=}4$); payload is driven by visual content rather than page count, so it cannot be
bounded by limiting pages. Within image control, the largest cells fit only at a lower
resolution (longest side 1024\,px at JPEG quality 80) than the rest (1536\,px at quality
90), so image resolution is mixed and correlated with document length, and we leave a
single-resolution re-render to an extension.

\subsection{Dataset and run statistics}
\label{app:stats}

This appendix gives the per-condition size and outcome statistics behind the main
results. Table~\ref{tab:dataset_stats} reports the size of each context condition: median
markdown characters, median pages or page-images, and the per-document page ranges for
sources and distractors. Table~\ref{tab:provisional} reports, for every model and
modality we ran, judge accuracy (the primary metric),
F1, and exact match.


\begin{table}[h]
\centering
\small
\begin{tabular}{lrrrr}
\toprule
\textbf{Condition} & \textbf{Markdown chars} & \textbf{Pages / images} & \multicolumn{2}{c}{\textbf{Per-doc pages}} \\
 & (median) & (median) & source & distractor \\
\midrule
Control   & 115{,}943 & 35  & \multirow{5}{*}{9 to 198 (med.\ 28)} & \multirow{5}{*}{10 to 40 (med.\ 23)} \\
HN-2      & 278{,}316 & 84  & & \\
Random-2  & 250{,}677 & 86  & & \\
HN-4      & 382{,}850 & 123 & & \\
Random-4  & 362{,}197 & 135 & & \\
\bottomrule
\end{tabular}
\caption{Per-cell size by condition (medians), and per-document page ranges.
Source documents are unbounded in length (9 to 198 pages); hard-negative and random
distractors are bounded to $[10,40]$ pages so that distractor type is not confounded
with length at a given $k$. Pages and page-images are one-to-one.}
\label{tab:dataset_stats}
\end{table}

\begin{table}[h]
\centering
\small
\begin{tabular}{lllrrr}
\toprule
\textbf{Model} & \textbf{Modality} & \textbf{Condition} & \textbf{Judge} & \textbf{F1} & \textbf{EM} \\
\midrule
\multicolumn{6}{l}{\emph{Markdown (all five conditions):}} \\
gpt-5-mini       & markdown & Control  & 0.763 & 0.666 & 0.448 \\
gpt-5-mini       & markdown & HN-2     & 0.737 & 0.641 & 0.444 \\
gpt-5-mini       & markdown & Random-2 & 0.767 & 0.659 & 0.444 \\
gpt-5-mini       & markdown & HN-4     & 0.707 & 0.632 & 0.448 \\
gpt-5-mini       & markdown & Random-4 & 0.748 & 0.639 & 0.437 \\
gemini-3.5-flash & markdown & Control  & 0.733 & 0.579 & 0.374 \\
gemini-3.5-flash & markdown & HN-2     & 0.711 & 0.545 & 0.352 \\
gemini-3.5-flash & markdown & Random-2 & 0.722 & 0.586 & 0.415 \\
gemini-3.5-flash & markdown & HN-4     & 0.726 & 0.561 & 0.385 \\
gemini-3.5-flash & markdown & Random-4 & 0.711 & 0.575 & 0.393 \\
grok-4.3         & markdown & Control  & 0.744 & 0.395 & 0.248 \\
grok-4.3         & markdown & HN-2     & 0.733 & 0.314 & 0.181 \\
grok-4.3         & markdown & Random-2 & 0.726 & 0.412 & 0.270 \\
grok-4.3         & markdown & HN-4     & 0.682 & 0.295 & 0.203 \\
grok-4.3         & markdown & Random-4 & 0.707 & 0.328 & 0.208 \\
\midrule
\multicolumn{6}{l}{\emph{Image and PDF (control only; distractor conditions infeasible, see \S\ref{sec:limitations}):}} \\
gpt-5-mini            & pdf   & Control & 0.804 & 0.701 & 0.493 \\
gemini-3.5-flash      & pdf   & Control & 0.778 & 0.651 & 0.448 \\
grok-4.3              & pdf   & Control & 0.667 & 0.271 & 0.152 \\
gpt-5-mini            & image & Control & 0.804 & 0.709 & 0.519 \\
gemini-3.5-flash      & image & Control & 0.789 & 0.655 & 0.470 \\
grok-4.3              & image & Control & 0.814 & 0.238 & 0.123 \\
\bottomrule
\end{tabular}
\caption{Per-condition judge accuracy (primary), F1, and EM. Markdown reports all five conditions for all three models. Image and PDF report control only; their distractor conditions are not run (Section~\ref{sec:limitations}).
\texttt{grok-4.3}'s F1 and EM are deflated by verbose answers (mean 254 response
characters against 34 for \texttt{gpt-5-mini}), not by lower correctness, so judge accuracy is the trustworthy metric for it.}
\label{tab:provisional}
\end{table}
\subsection{Prompts}
\label{app:prompts}

\paragraph{Model instruction (markdown).}
\begin{quote}\small\ttfamily
You are given one or more documents (converted to markdown). Using ONLY their
content, answer the question with a short, direct answer and nothing else. If the
documents do not contain the answer, reply "Not answerable".
\end{quote}
The image and PDF variants replace ``documents (converted to markdown)'' with
``documents as page images'' and ``PDF documents'' respectively.

\paragraph{Judge instruction.}
\begin{quote}\small\ttfamily
You are scoring a model's answer against a reference answer for a long-context
document QA benchmark. [question, reference, prediction inserted] Rules: mark
correct: true only if the prediction conveys the same factual content as the
reference; numbers and units are matched by value; list order does not matter and
missing items earn partial credit; minor paraphrase is acceptable; refusals are
incorrect. Return only a JSON object \{"correct": true|false, "score": 0.0-1.0,
"reasoning": "one short sentence"\}. 
\end{quote}

\end{document}